\documentclass[12pt,a4paper]{article}

\usepackage[cmex10,intlimits]{amsmath}
\usepackage{color}
\usepackage{a4wide}
\usepackage{rotating}
\usepackage{cite}
\usepackage{tikz}
\usepackage{bm}
\usepackage{tipa}
\usepackage[utf8]{inputenc}
\usepackage[T1]{fontenc}

\usepackage{hyperref}       
\usepackage{url}            
\usepackage{booktabs}       
\usepackage{amsfonts}       
\usepackage{nicefrac}       
\usepackage{microtype}      
\usepackage[intlimits]{amsmath}
\usepackage{amssymb}
\usepackage{mathptmx}

\usepackage{subcaption}
\usepackage{graphicx}

\makeatletter
\def\blfootnote{\xdef\@thefnmark{}\@footnotetext}
\makeatother

\title{A Brief Prehistory of Double Descent\footnote{PNAS MS\# 2020-01875 Letter to Editor.  Accepted for electronic publications in the Proceedings of the National Academy of Sciences of the United States of America (\url{https://www.pnas.org/}).}}
\author{Marco~Loog$^{1,2}$ \, Tom~Viering$^2$ \,  Alexander~Mey$^2$ \\  Jesse~H.~Krijthe$^2$ \, David~M.J.~Tax$^2$ \medskip \small \\
\small
\begin{tabular}{c}
$^1$University of Copenhagen, Denmark \\
$^2$Delft University of Technology, The Netherlands
\end{tabular}}

\begin{document}

\maketitle



In their thought-provoking paper \cite{belkin2019reconciling}, Belkin et al.\ illustrate and discuss the shape of risk curves in the context of modern high-complexity learners. Given a fixed training sample size $n$, such curves show the risk of a learner as a function of some (approximate) measure of its complexity $N$.  With $N$ the number of features, these curves are also referred to as feature curves.  A salient observation in \cite{belkin2019reconciling} is that these curves can display, what they call, double descent: with increasing $N$, the risk initially decreases, attains a minimum, and then increases until $N$ equals $n$, where the training data is fitted perfectly.  Increasing $N$ even further, the risk decreases a second and final time, creating a peak at $N=n$.  This twofold descent may come as a surprise, but as opposed to what \cite{belkin2019reconciling} reports, it has not been overlooked historically. Our letter draws attention to some original, earlier findings, of interest to contemporary machine learning.

Already in 1989, using artificial data, Vallet et al.\ \cite{vallet1989linear} experimentally demonstrate double descent for \emph{learning curves} of classifiers trained through minimum norm linear regression (MNLR, see \cite{penrose1956best})---termed the pseudo-inverse solution in \cite{vallet1989linear}.  In learning curves the risk is displayed as a function of $n$, as opposed to $N$ for risk curves. What intuitively matters in learning behavior, however, is the sample size \emph{relative} to the measure of complexity.  This idea is made explicit in various physics papers on learning (e.g. \cite{vallet1989linear,opper1990ability,watkin1993statistical}), where the risk is often plotted against $\alpha = \tfrac{n}{N}$.  A first theoretical results on double descent, indeed using such $\alpha$, is given by Opper et al.\ \cite{opper1990ability}. They proof that in particular settings, for $N$ going to infinity, the pseudo-inverse solution improves as soon as one moves away from the peak at $\alpha=1$.

Employing a so-called pseudo-Fisher linear discriminant (PFLD, equivalent to MNLR), Duin \cite{duin2000classifiers} is the first to show feature curves on real-world data quite similar to the double-descent curves in \cite{belkin2019reconciling}.  Compare, for instance, Fig.\ 2 in \cite{belkin2019reconciling} with Fig.\ 6 and 7 from \cite{duin2000classifiers}.  Skurichina and Duin \cite{skurichina1998regularization} demonstrate experimentally that increasing PFLD's complexity simply by adding random features can improve performance when $N=n$ (i.e., $\alpha = 1$).  The benefit of some form of regularization has been shown already in \cite{vallet1989linear}.  For semi-supervised PFLD, Krijthe and Loog \cite{krijthe2016peaking} demonstrate that unlabeled data can regularize, but also worsen the peak in the curve.  Their work builds on the original analysis of double descent for the supervised PFLD by Raudys and Duin \cite{raudys1998expected}.

Interestingly, results from \cite{opper1990ability,watkin1993statistical,duin2000classifiers,skurichina1998regularization} suggest that some losses may not exhibit double descent in the first place.  In \cite{duin2000classifiers,skurichina1998regularization}, the linear SVM shows regular monotonic behavior. Analytic results from \cite{opper1990ability,watkin1993statistical} show the same for the so-called perceptron of optimal (or maximal) stability, which is closely related to the SVM \cite{watkin1993statistical}.

The findings in \cite{belkin2019reconciling} go, significantly, beyond those for the MNLR. Also shown, for instance, is double descent for 2-layer neural networks and random forests.  Combining this with observations such as those from Loog et al.\ \cite{loog2019minimizers}, which show striking multiple-descent learning curves (even in the underparameterized regime), the need to further our understanding of such rudimentary learning behavior is evident.

\bibliography{pnas-sample}
\bibliographystyle{unsrt}


\end{document}